\documentclass[letterpaper, 10 pt, conference]{ieeeconf}
\usepackage{booktabs}
\usepackage{bm}
\usepackage{pifont}
\usepackage{amsmath}
\usepackage{graphicx}
\usepackage[export]{adjustbox}
\usepackage{multirow}
\usepackage{array}
\usepackage{pifont}
\newcolumntype{M}[1]{>{\centering\arraybackslash}p{#1}}
\usepackage{epsfig}
\usepackage{tabularray}
\usepackage{verbatim}
\usepackage{times}
\usepackage{xcolor}
\usepackage{cite}
\usepackage{subfigure}
\usepackage{amsmath}
\usepackage{color}
\usepackage{url}
\usepackage{amssymb}
\usepackage{marvosym}
\usepackage{ulem}
\usepackage[colorlinks=true, urlcolor=blue, linkcolor=red]{hyperref}
\hypersetup{
    colorlinks=true,
    linkcolor=blue,
    filecolor=black,      
    urlcolor=purple,
    citecolor=red
}

\IEEEoverridecommandlockouts
\title{\LARGE \bf
Generalized Correspondence Matching via Flexible Hierarchical Refinement and Patch Descriptor Distillation
}

\author{ \ \ Yu Han, Ziwei Long, Yanting Zhang\textsuperscript{\Letter}, Jin Wu, Zhijun Fang, Rui Fan\textsuperscript{\Letter}
\thanks{
This research was supported by the National Natural Science Foundation of China under Grants 62233013, 62206046, and U2033218, the Shanghai Sailing Program under Grant 21YF1401300, the Science and Technology Commission of Shanghai Municipal under Grant 22511104500, the Fundamental Research Funds for the Central Universities, and Xiaomi Young Talents Program. (\emph{Y. Han and Z. Long contributed equally to this work}) ({\Letter} \emph{Corresponding authors: Y. Zhang and R. Fan}).
}
\thanks{
Y. Han, Y. Zhang, and Z. Fang are with the School of Computer Science and Technology, Donghua University, Shanghai 201620, P. R. China
(e-mails: {2232816@mail.dhu.edu.cn, \{ytzhang, zjfang\}@dhu.edu.cn}).
}
\thanks{
J. Wu is with the Department of Electronics and Computer Engineering, the Hong Kong University of Science and Technology, Clear Water Bay, Hong Kong SAR, P. R. China 
(e-mail: jin\_wu\_uestc@hotmail.com).
}
\thanks{
Z. Long and R. Fan are with the Machine Intelligence \& Autonomous Systems (MIAS) Group, the College of Electronics \& Information Engineering, Shanghai Research Institute for Intelligent Autonomous Systems, the State Key Laboratory of Intelligent Autonomous Systems, and Frontiers Science Center for Intelligent Autonomous Systems, Tongji University, Shanghai 201804, P. R. China (e-mails: \{2053301, rfan\}@tongji.edu.cn).
}
}
\begin{document}

\maketitle
\thispagestyle{empty}
\pagestyle{empty}

\begin{abstract}
Correspondence matching plays a crucial role in numerous robotics applications. In comparison to conventional hand-crafted methods and recent data-driven approaches, there is significant interest in plug-and-play algorithms that make full use of pre-trained backbone networks for multi-scale feature extraction and leverage hierarchical refinement strategies to generate matched correspondences. The primary focus of this paper is to address the limitations of deep feature matching (DFM), a state-of-the-art (SoTA) plug-and-play correspondence matching approach. First, we eliminate the pre-defined threshold employed in the hierarchical refinement process of DFM by leveraging a more flexible nearest neighbor search strategy, thereby preventing the exclusion of repetitive yet valid matches during the early stages. Our second technical contribution is the integration of a patch descriptor, which extends the applicability of DFM to accommodate a wide range of backbone networks pre-trained across diverse computer vision tasks, including image classification, semantic segmentation, and stereo matching. Taking into account the practical applicability of our method in real-world robotics applications, we also propose a novel patch descriptor distillation strategy to further reduce the computational complexity of correspondence matching. Extensive experiments conducted on three public datasets demonstrate the superior performance of our proposed method. Specifically, it achieves an overall performance in terms of mean matching accuracy of 0.68, 0.92, and 0.95 with respect to the tolerances of 1, 3, and 5 pixels, respectively, on the HPatches dataset, outperforming all other SoTA algorithms. Our source code, demo video, and supplement are publicly available at \url{mias.group/GCM}. 
\end{abstract}

\section{Introduction}
\label{sec:introduction}

\begin{figure}[t!]
      \centering
      \includegraphics[width=0.48\textwidth]{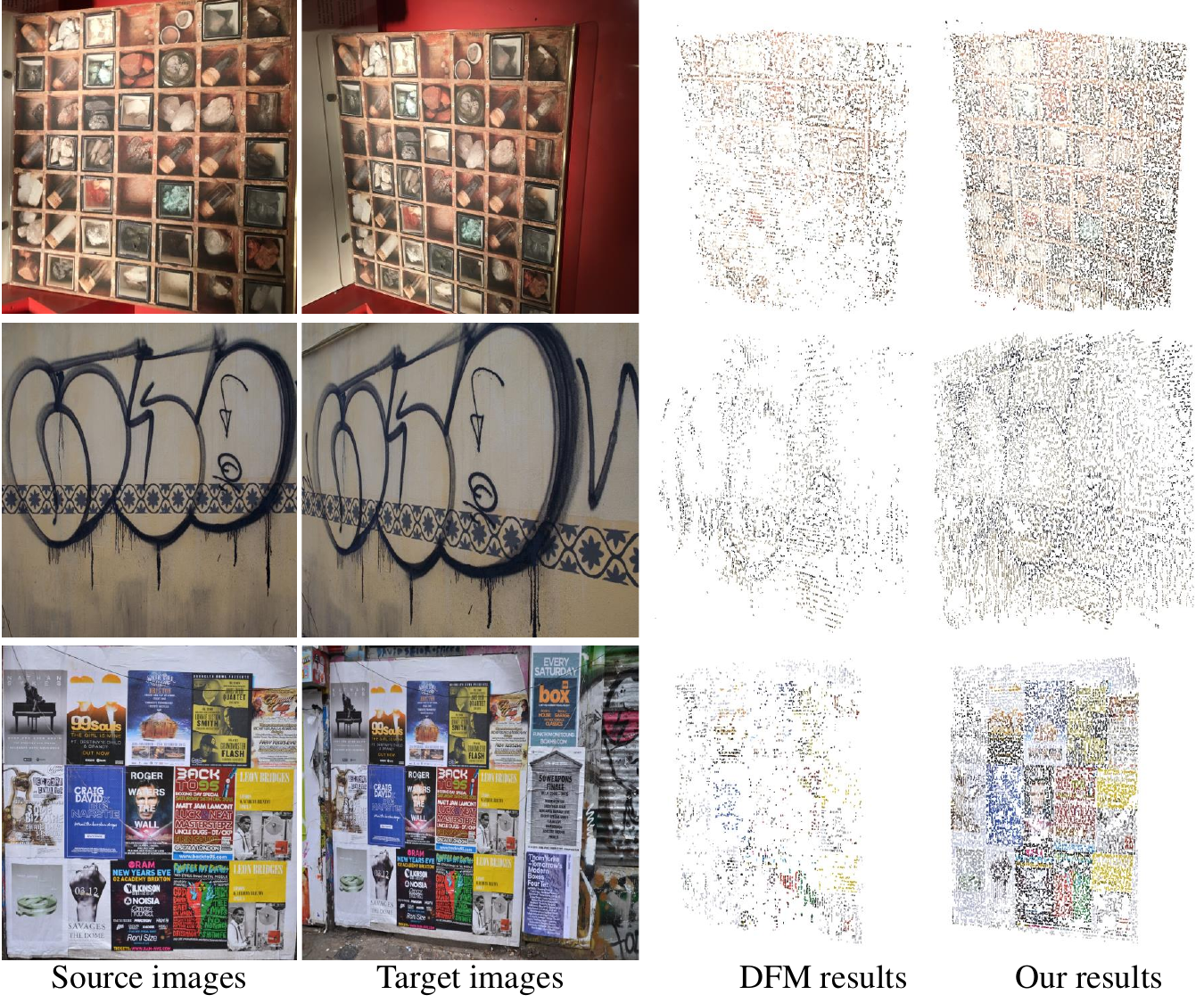}  
       \vspace{-2em}
      \caption{
      \textbf{Comparison between DFM and our proposed GCM on the Hpatches dataset.} Our approach produces more correspondences in the regions with low texture or repetitive patterns. 
      }
      \label{fig.cover}
      \vspace{-2.0em}
\end{figure}

Correspondence matching between images is crucial for a wide range of computer vision and robotics applications, \textit{e.g.}, simultaneous localization and mapping \cite{bhutta2020loop, 7219438, bhutta2018pcr}, 3D geometry reconstruction \cite{fan2018road, SFM, fan2021rethinking}, and stereo matching \cite{wu2024s3mnet, zhao2023dive, wang2021pvstereo, wu2024icra}. Conventional hand-crafted approaches extract keypoints using human-designed local feature detectors and descriptors, such as the scale-invariant feature transform (SIFT) \cite{SIFT} and speeded up robust features (SURF) \cite{bay2006surf}. Correspondence pairs are then determined using the nearest neighbor search (NNS) algorithm \cite{NN}. With the recent advances in deep learning, data-driven approaches \cite{detone2018superpoint,dusmanu2019d2,NEURIPS2019_R2D2,sarlin2020superglue,rocco2020ncnet,rocco2020efficient,sun2021loftr} have demonstrated compelling results. 

Conventional hand-crafted approaches generally leverage a sequential pipeline for keypoint detection, description, and matching \cite{bhutta2021towards, ma2021image}. Their overall performance is often determined by the weakest component within this pipeline, akin to the ``barrel effect'' principle. Moreover, errors in the earlier stages can accumulate and propagate to the later stages, making it tricky to improve the overall performance \cite{IME}. While data-driven approaches have significantly outperformed hand-crafted methods, detector-based methods \cite{detone2018superpoint,dusmanu2019d2,NEURIPS2019_R2D2,sarlin2020superglue} may still struggle in texture-less regions, and detector-free approaches could face information loss due to manually selected scales  \cite{rocco2020ncnet,rocco2020efficient,sun2021loftr}. Additionally, most data-driven approaches require a large amount of well-annotated data for model training, often resulting in unsatisfactory performance when applied to new scenarios \cite{zhou2023e3cm}.

To address these limitations, deep feature matching (DFM) \cite{dfm}, a plug-and-play approach built upon a hierarchical matching refinement paradigm is proposed. DFM utilizes a VGG \cite{simonyan2014very} model pre-trained on the ImageNet \cite{NIPS2012_c399862d} database to extract multi-scale features, with no need for additional training with well-annotated data. Furthermore, DFM leverages a coarse-to-fine strategy in conjunction with the NNS algorithm to perform hierarchical feature matching from the deepest layers to the shallowest ones. DFM significantly improves accuracy and robustness compared to conventional hand-crafted methods, outperforming even some approaches trained with correspondences \cite{zhou2023e3cm}. 

Nevertheless, DFM has three significant limitations. The most fatal drawback is its high demand on the backbone network, limiting compatibility to those capable of providing feature maps with the same size as the input image, such as VGG. Another drawback is the lack of extensive experimental evaluation of DFM in various computer vision and robotics tasks, except for image classification. This raises questions about its performance when using different backbone networks, especially those trained for dense correspondence matching, such as \cite{duggal2019deeppruner,li2022practical,chang2018pyramid}. Finally, the hierarchical refinement strategy employed in DFM has the potential to propagate errors from deeper layers to shallower ones, resulting in lower density and quality of correspondence matching, as shown in Fig. \ref{fig.cover}.

To address the challenges mentioned above, this paper introduces \uline{\textbf{G}eneralized \textbf{C}orrespondence \textbf{M}atching (\textbf{GCM})} based on flexible hierarchical refinement and patch descriptor distillation. First, we omit the pre-defined threshold used in the hierarchical refinement process of DFM by leveraging a more flexible NNS strategy, thereby preventing the exclusion of repetitive yet valid matches in early stages. Furthermore, we expand the applicability of DFM to accommodate various types of backbones, pre-trained across diverse computer vision tasks, including image classification \cite{simonyan2014very, he2016deep, xie2017aggregated, howard2019searching, tan2021efficientnetv2, ma2018shufflenet}, semantic segmentation \cite{ronneberger2015u, chen2017rethinking, long2015fully, liu2021swin}, and stereo matching \cite{duggal2019deeppruner, chang2018pyramid, li2022practical}. This is accomplished by incorporating a patch descriptor to function as the highest-resolution feature maps (with the same resolution as the input image) in DFM. Additionally, we propose a novel strategy for patch descriptor distillation, which further enhances the overall efficiency of correspondence matching. What surprises us is that several backbones demonstrate improved performance when the patch descriptor is distilled. Extensive experiments conducted on the HPatches dataset\cite{balntas2017hpatches} demonstrate the superior mean matching accuracy (MMA) achieved by GCM. Moreover, as illustrated in Fig. \ref{fig.cover}, GCM is capable of producing denser and more accurate matched correspondences, particularly in repetitive or low-texture areas when compared to DFM.

\begin{figure*}[t!]
      \centering
      \includegraphics[width=0.999\textwidth]{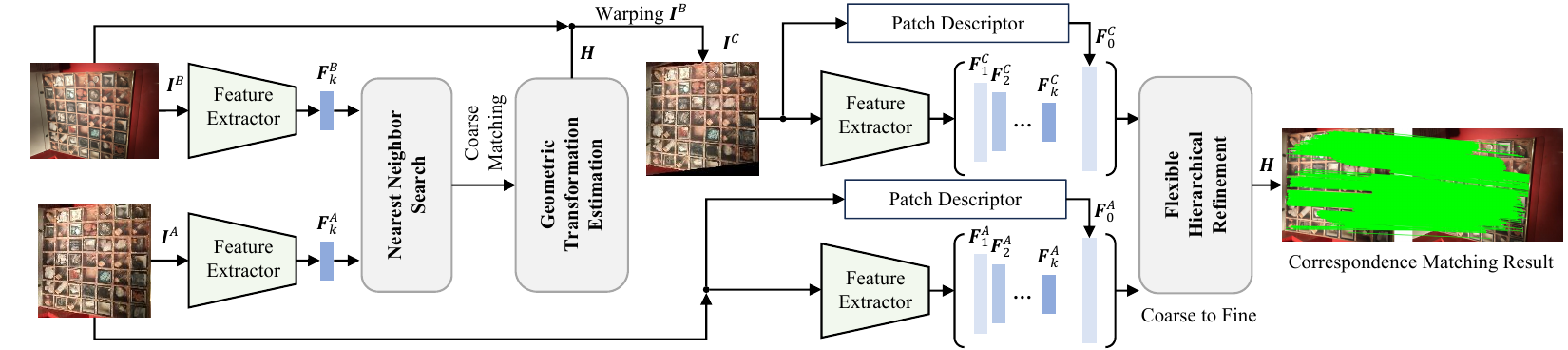}
\vspace{-1.5em}
      \caption{\textbf{The architecture of our proposed GCM.}}
      \label{fig.overview}
\vspace{-1.6em}
\end{figure*}

\section{Related Work}
\label{sec:related_work}

\subsection{Correspondence Matching}

Early approaches, as exemplified by SIFT \cite{SIFT}, typically employ hand-crafted local visual features, \textit{e.g.}, gradients, angles, and blobs, to detect distinctive interest points and generate descriptors. Correspondence pairs are subsequently determined via NNS \cite{zhou2023e3cm}. However, these algorithms have limited adaptability to diverse and complex datasets owing to their reliance on manually designed features \cite{NEURIPS2019_R2D2}. Moreover, they tend to be highly sensitive to variations in lighting, scale, and viewpoint, which constrains their robustness in real-world scenarios \cite{detone2018superpoint}. In addition, they often require significant domain expertise for feature design and may struggle to generalize effectively across different tasks or domains.

Over the last half-decade, data-driven methods \cite{NEURIPS2019_R2D2,dusmanu2019d2, sarlin2020superglue} have demonstrated superior performance compared to conventional hand-crafted approaches. Among them, detector-based algorithms have remained the dominant choice for correspondence matching. As an example, SuperPoint \cite{detone2018superpoint} leverages homographic adaptation in conjunction with MagicPoint \cite{detone2017toward} to enhance detector performance and generate pseudo ground-truth interest points for unlabeled images in a self-supervised fashion. Another data-driven approach, repeatable and reliable detector and descriptor (R2D2) \cite{NEURIPS2019_R2D2}, is built upon a trainable convolutional neural network (CNN). R2D2 enhances the descriptor quality by placing a strong emphasis on the reliability of feature points. This is achieved in conjunction with a trainable CNN designed for both feature description and detection, known as D2-Net \cite{dusmanu2019d2}. Such improvements greatly advance R2D2's feature description capabilities. In contrast, DFM \cite{dfm}, which relies on VGG-19 model, does not offer such precision in the description. Additionally, SuperGlue \cite{sarlin2020superglue} leverages a graph neural network (GNN) equipped with an attention mechanism and a differentiable Sinkhorn algorithm \cite{sinkhorn1964relationship} to compute matches between two sets of detected features and their corresponding descriptor vectors. SuperGlue achieves notably superior performance when compared to the NNS methods.

Other approaches, such as neighborhood consensus networks (NCNet) designed for image correspondence estimation \cite{rocco2020ncnet} forgo the feature detection phase and instead directly match points distributed across a dense grid rather than sparse locations. NCNet constructs a 4D cost volume with neighborhood consensus to enumerate all possible matches between images. In contrast, efficient neighborhood consensus networks via submanifold sparse convolutions (SparseNCNet) \cite{rocco2020efficient} employs a more condensed form of the correlation tensor, storing a subset and substituting the dense 4D convolution with a sparser convolution technique to improve correspondence matching efficiency. Epipolar-guided pixel-level correspondences (Patch2pix) \cite{zhou2021patch2pix} utilizes a pre-trained backbone to extract patch-level matches and employs a two-stage regressor to refine these matches to pixel-level precision. On the other hand, detector-free local feature matching with Transformers (LoFTR) \cite{sun2021loftr} achieves a high level of robustness but can be slower due to the large number of processes it involves. Although efficiency can be improved by reducing the resolution of the input image, this may affect matching accuracy to some extent. In this paper, our GCM is built upon the foundation of DFM. Our primary focus is to tackle the limitations of DFM, particularly the high demand for feature maps and the necessity of a predefined threshold for hierarchical refinement.

\subsection{Knowledge Distillation}

Knowledge distillation is a commonly used technique for model compression, initially introduced for image classification. Unlike pruning and quantization techniques \cite{park2022quantized} used in model compression, knowledge distillation techniques focus on training a smaller, more lightweight model \cite{zhao2022decoupled}. This is achieved by exploiting supervised information from larger, high-performance models, ultimately reducing both time and space complexity. Due to its success across various tasks, knowledge distillation is regarded as an effective multitasking approach, applicable to classification \cite{hinton2015distilling}, semantic segmentation \cite{he2019knowledge}, and object detection \cite{li2017mimicking, chen2017learning}. In this paper, we utilize the knowledge distillation technique to further reduce the complexity of the patch descriptor.


\section{Methodology}
\label{sec:methods}

\subsection{Architecture Overview}


The architecture of our proposed GCM is shown in Fig. \ref{fig.overview}. GCM follows a two-step correspondence matching strategy, similar to DFM \cite{dfm}. Initially, we employ a pre-trained backbone network as the deep feature extractor to obtain the feature maps $\mathcal{F}=\{\bm{F}_1,...,\bm{F}_k\}$. Here, $\boldsymbol{F}_l \in \mathbb{R}^{\frac{H}{2^l} \times \frac{W}{2^l} \times C_l}$ represents the $l$-th layer of feature maps, with $H$ and $W$ denoting the height and width of the input RGB image $\bm{I} \in \mathbb{R}^{H \times W \times  3}$, respectively. Next, we perform NNS on the last layers of feature maps, denoted as $\bm{F}_k^A$ and $\bm{F}_k^B$, to obtain coarse matches $\mathcal{M}_k^{A,B}$. Here, the superscripts $A$ and $B$ represent two images of $\bm{I}^A$ and $\bm{I}^B$. Based on these matches, we estimate the homography matrix $\bm{H}$, which is further utilized to generate a warped image $\bm{I}^C$ from $\bm{I}^B$.

In the second step, we perform feature extraction followed by a flexible hierarchical refinement (FHR) module, which will be detailed in Section \ref{sec:Hierarchical}, to obtain the matched correspondences. In specific, we construct hierarchical feature map layers $\{\bm{F}_1^A,\bm{F}_2^A,...,\bm{F}_k^A, \bm{F}_0^A\}$ for $\bm{I}^A$, and $\{\bm{F}_1^B,\bm{F}_2^B,...,\bm{F}_k^B, \bm{F}_0^B\}$ for $\bm{I}^B$. Note that, $\bm{F}_0 \in \mathbb{R}^{H \times W \times C_0}$ is obtained through a patch descriptor described in Section \ref{sec:KL}, which provides feature maps with the same size as the original images. The FHR module operates in a coarse-to-fine manner, processing from the deepest layer $\bm{F}_k$ to shallow layers down to $\bm{F}_1$, and finally $\bm{F}_0$, where we obtain the final matches between $\bm{I}^A$ and the warped image $\bm{I}^C$. Later, taking advantage of the estimated $\bm{H}$ in the first step, we can trace back to the matched correspondences between $\bm{I}^A$ and the original image of $\bm{I}^B$.

Our method generalizes the baseline DFM approach and is compatible with various types of backbone networks. These networks can be pre-trained for a diverse range of computer vision tasks, regardless of whether they can produce feature maps with the same resolution as the original images.

\subsection{Nearest Neighbor Search}
\label{sec:Hierarchical}

Since the dense NNS utilized in DFM is a local matching approach that necessitates a manually defined threshold and tends to discard correct matches in cases of repetitiveness, we adopt a flexible, parameter-free NNS strategy.

Given the feature maps $\bm{F}^A$ and $\bm{F}^B$ extracted from images $\bm{I}^A$ and $\bm{I}^B$, we identify potential matches by determining the nearest neighbors based on the feature distance between $p^A$ and $p^B$ (with $\bm{f}^A$ and $\bm{f}^B$ denoting the features at the point $p^A$ and $p^B$ in $\bm{F}^A$ and $\bm{F}^B$, respectively):
\begin{equation}
d(p^A ,p^B)=1-\phi(\bm{f}^A, \bm{f}^B),
\label{eq:distance}
\end{equation}
where $\phi(\cdot,\cdot)$ represents the cosine similarity as follows:
\begin{equation}
    \phi(\bm{f}^A, \bm{f}^B) = \frac{\bm{f}^A \cdot \bm{f}^B }{\left \| \bm{f}^A \right \|_{2} \left \| \bm{f}^B  \right \|_{2} }.
    \label{eq:phi}
\end{equation}
For a specific point $p^A$ within the feature map $\bm{F}^A$, it is matched to $p^B$ if the distance to $p^B$ is minimal. A match $(p^A, p^B)$ is confirmed only if it is mutual, meaning that $p^A$ and $p^B$ are recognized as a matched pair only if $p^B$ is also matched with $p^A$.
\subsection{Patch Descriptor Distillation} 
\label{sec:KL}
\begin{figure}[t!]
\vspace{+0.51em}
      \centering
      \includegraphics[width=0.38\textwidth]{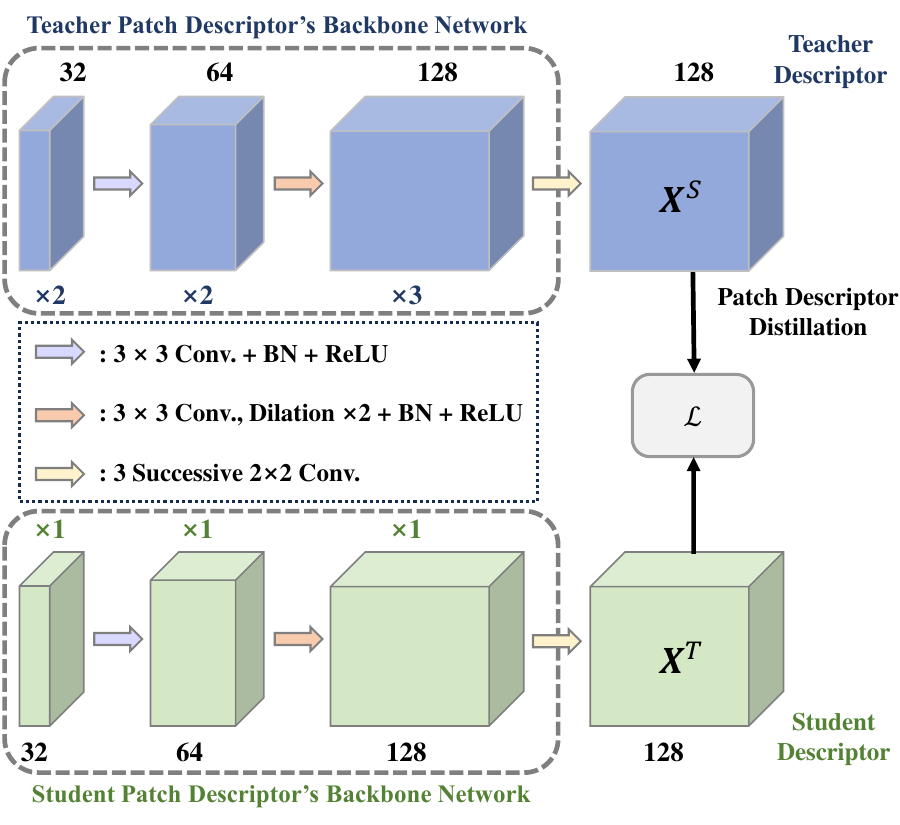}
\vspace{-1em}
      \caption{\textbf{Our proposed patch descriptor distillation strategy}. We employ the R2D2 backbone as the teacher descriptor to generate a more lightweight student descriptor. BN refers to batch normalization.}
      \label{fig:R2D2_distillation}
\vspace{-1.5em}
\end{figure}

To reduce time and space complexity, we develop an additional distilled patch descriptor as a more lightweight feature description network alternative. Given that descriptors of R2D2 \cite{NEURIPS2019_R2D2} are sourced from an L2-normalized feature map, our strategy is to train the final layer of the student model's backbone using the final feature layer of the teacher model's backbone. As shown in Fig. \ref{fig:R2D2_distillation}, the student model reduces the number of intermediate layers and directly assimilates the final feature map from the teacher model. $\boldsymbol{X}^{T} \in \mathbb{R}^{H \times W \times 128}$ denotes the output of the last layer of the teacher model backbone network, and $\boldsymbol{X}^{S} \in \mathbb{R}^{H \times W \times 128}$ denotes the last layer output of the student model backbone network. The loss function is defined as follows:
\begin{equation}
     \mathcal{L} = \frac{1}{|\mathcal{P}|} \sum_{(i, j) \in \mathcal{P}}(1-|\phi(\boldsymbol{X}^{T}_{ij}, \boldsymbol{X}^{S}_{ij})|),
     \label{eq:distillation}
\end{equation}
where the point set $\mathcal{P}$ contains all points in the image, represented by coordinates $(i, j)$ for each pixel. The function $\phi$ is defined in Equation (\ref{eq:phi}). This lightweight patch descriptor enables the faster generation of $\boldsymbol{F}_0 \in \mathbb{R}^{H \times W \times 128}$ for facilitating flexible hierarchical refinement.

\subsection{Flexible Hierarchical Refinement}
\label{sec:Hierarchical}
\begin{figure}[t!]
\vspace{+0.5em}
      \centering
      \includegraphics[width=0.475\textwidth]{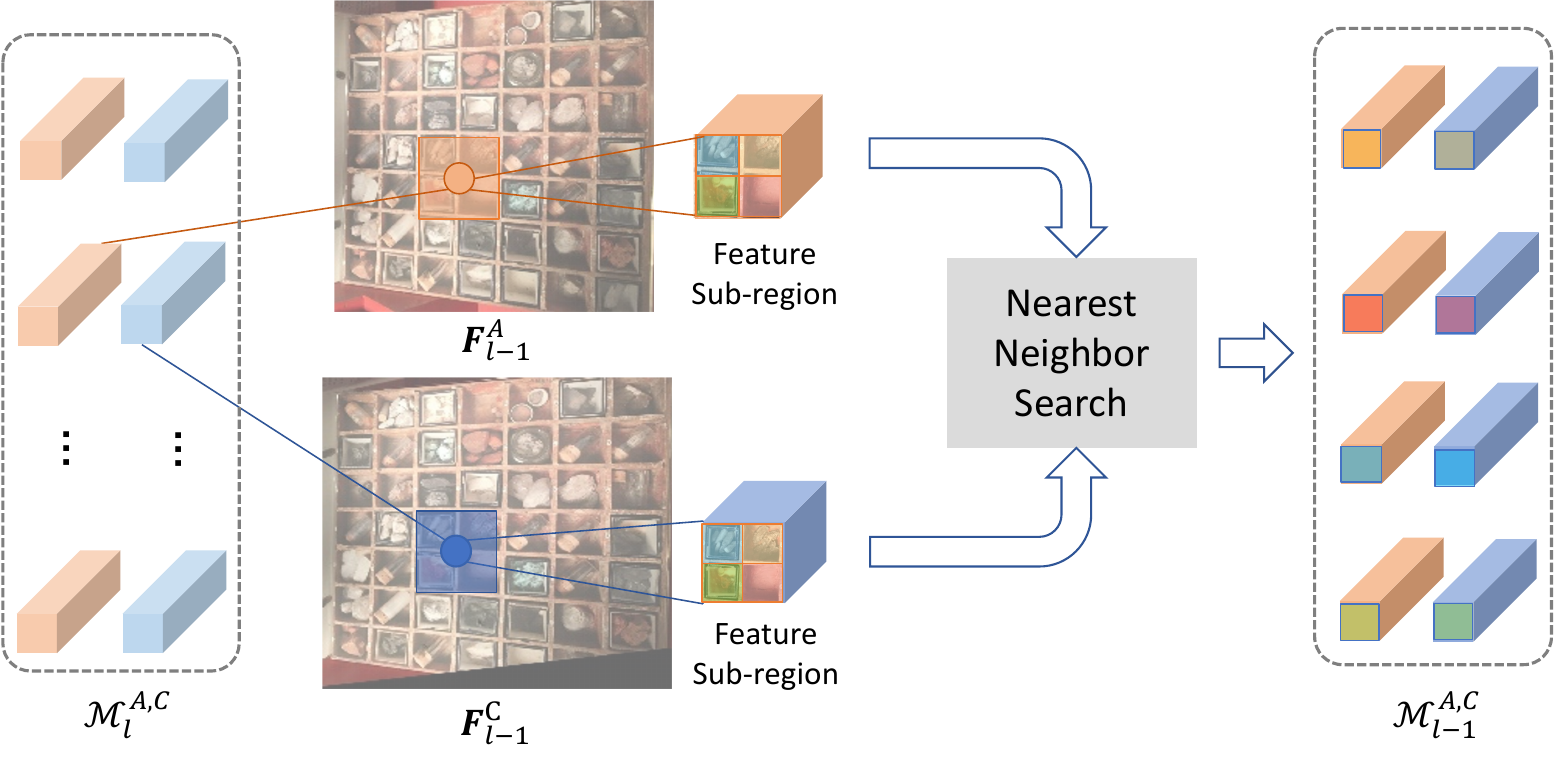}   
      \vspace{-0.8em}
      \caption{\textbf{An illustration of one iteration within the hierarchical refinement module.}
      }
      \label{fig.fhr}
      \vspace{-1.5em}
\end{figure}

The flexible hierarchical refinement (FHR) module consists of multiple iterations, with a single iteration illustrated in Fig. \ref{fig.fhr}. Given the matched features $\mathcal{M}_l^{A,C}$ in the $l$-th layer, 
where every match $( p_l^A, p_l^C) \in \mathcal{M}_l^{A,C}$ is associated with feature $(\bm{f}_l^A, \bm{f}_l^C)$, this module returns matched correspondences $\mathcal{M}_{l-1}^{A,C}$ in the $(l-1)$-th layer. 
During this step, it allows the matched features in $\bm{F}_l^A$ and $\bm{F}_l^C$ to identify the corresponding sub-region pairs in $\bm{F}_{l-1}^A$ and $\bm{F}_{l-1}^C$. In specific, every point in $\bm{F}_{l}$ is mapped to $\bm{F}_{l-1}$ and corresponds to a $2 \times 2$ patch on $\bm{F}_{l-1}$. 
By applying NNS to these sub-regions, matched correspondences $\mathcal{M}_{l-1}^{A,C}$ between $\bm{F}_{l-1}^A$ and $\bm{F}_{l-1}^C$ can be obtained.

There are $(k+1)$ iterations in total, consistent with the number of hierarchical feature layers from $\bm{F}_k$ to $\bm{F}_0$. As for the initial input of FHR, we employ the NNS on the $k$-th layer's $\bm{F}_k$ from the two images to obtain an initial correspondence. Subsequently, we map these points to $\bm{F}_{k-1}$ and establish paired sub-regions between $\bm{I}^A$ and $\bm{I}^C$.
Using this correspondence, we employ the NNS to establish refined correspondences within each group of patches. This process is iteratively applied until the first layer of $\bm{F}_1$ is reached, resulting in correspondences at half size of the input images. 
Finally, we project the matched points onto the descriptor layer $\bm{F}_0$ and perform correspondence matching at this layer using NNS between every paired patch associated with $\bm{I}^A$ and $\bm{I}^C$.
Due to potential accumulated errors from the previous stages, we introduce a ratio test as in \cite{dusmanu2019d2, SIFT} at the final iteration to filter out inferior matches. 

It is worth noting that this hierarchical refinement paradigm is adaptable to a wide range of backbone networks.

\section{EXPERIMENTS}
\label{sec:experiments}

\subsection{Datasets and Evaluation Protocol}
Here is a summary of the datasets and evaluation setups used in our experiments:

\ding{172} HPatches: We conduct experiments on the HPatches dataset, utilizing mean matching accuracy (MMA) and homography estimation accuracy as metrics. For both image matching and homography estimation, we follow the evaluation setup detailed in \cite{dfm}.

\ding{173} MegaDepth: We evaluate the outdoor pose estimation accuracy of our proposed method on the MegaDepth dataset \cite{li2018megadepth}. We quantify the pose error by computing the area under the receiver operating characteristic (AUC), following the setup detailed in \cite{tyszkiewicz2020disk} and \cite{sun2021loftr}. Additionally, we also compute match precision following the setup in \cite{sarlin2020superglue}.

\ding{174} ScanNet: We evaluate the indoor pose estimation accuracy on the ScanNet dataset \cite{dai2017scannet}. Please refer to \cite{sarlin2020superglue} for detailed setups.

\begin{table*}[t!]
    \vspace{+0.5em}
    \fontsize{7.5}{9.0}\selectfont
    \centering
    \caption{\textbf{MMA comparison on the HPatches dataset.} ``$r$" denotes ratio test threshold.}
    \vspace{-1em}
    \begin{tabular}[width=0.95\linewidth]{M{2.4cm}M{2.6cm}  ccc|ccc|ccc c}
\toprule
\multirow{2}{*}{Category}&\multirow{2}{*}{\centering Method}  & \multicolumn{3}{c|}{Overall} & \multicolumn{3}{c|}{Illumination} & \multicolumn{3}{c}{Viewpoint} & \multirow{2}{*}{Matches} \\
\cline{3-11}
\multicolumn{2}{c}{} & @1px & @3px & @5px & @1px & @3px & @5px & @1px & @3px & @5px & \\
\hline
Hand-Crafted&SIFT+NNS\cite{SIFT}           & 0.35 & 0.50 & 0.54    & 0.37 & 0.49 & 0.52    & 0.33 & 0.52 & 0.55      & 0.4K                   \\
 \hline
 \multirow{2}{*}{Fully Supervised}&R2D2+NNS\cite{NEURIPS2019_R2D2}         & 0.33 & 0.76 & 0.84    & 0.38 & 0.81 & 0.90    & 0.29 & 0.71 & 0.79      & 1.6k                    \\
&LoFTR\cite{sun2021loftr}          & 0.63 & 0.91 & 0.93 & 0.68 & \textbf{0.95} & 0.96  &  0.59 & 0.86 & 0.91 & 2.6K
\\
\hline

\multirow{4}{*}{Plug-and-Play} & DFM ($r$=0.90) ~\cite{dfm}   & 0.51 & 0.85 & 0.93    & 0.63 & 0.91 & 0.97    & 0.42 & 0.81 & 0.89      & 7.0K              \\
&DFM ($r$=0.60) ~\cite{dfm}   & 0.61 & 0.88 & 0.94    & \textbf{0.77} & 0.93 & \textbf{0.98}    & 0.47 & 0.84 & 0.90      & 1.0K                   \\
&Ours ($r$=0.95)  &0.53&0.85& 0.92  &0.56  &0.87  &0.94  &0.50  &0.84  &0.90     &\textbf{14.4K}     \\
 &Ours ($r$=0.60)                         &\textbf{0.68}&\textbf{0.92}&\textbf{0.95} &0.74&0.93&0.97 &\textbf{0.63}&\textbf{0.91}&\textbf{0.94}    &3.1K  \\
\bottomrule
\end{tabular}

    \label{tab:exp_MMA}
    \vspace{-0.8em}
\end{table*}
\begin{table*}[t!]
    \centering
    \fontsize{7.5}{9.0}\selectfont
    \caption{\textbf{MMA comparison on the HPatches dataset among backbone networks pre-trained for a variety of computer vision tasks.} ``$r$" denotes ratio test threshold. ``$\ast$'' denotes the backbone used in R2D2 \cite{NEURIPS2019_R2D2}.}
    \vspace{-1em}
    
\begin{tabular}[width=0.999\linewidth]{M{2.4cm}M{2.6cm} ccc|ccc|ccc c}

\toprule
\multirow{2}{*}{Task}&\multirow{2}{*}{Method} & \multicolumn{3}{c|}{Overall} & \multicolumn{3}{c|}{Illumination} & \multicolumn{3}{c}{Viewpoint} & \multirow{2}{*}{Matches} \\
\cline{3-11}
\multicolumn{2}{c}{} & @1px & @3px & @5px & @1px & @3px & @5px & @1px & @3px & @5px & \\
\hline
  \multirow{7}{2.4cm}{\centering Image\\ Classification}& ResNet18 \cite{he2016deep}  &0.53&0.85& 0.92  &0.56  &0.87  &0.94  &0.50  &0.84  &0.90     &14.4K     \\
 &ResNet18 ($r$=0.6) \cite{he2016deep}                         &0.68&\textbf{0.92}&\textbf{0.95} &0.74&0.93&0.97 &0.63&0.91&\textbf{0.94}    &3.1K  \\
 &ResNet18* ($r$=0.6) \cite{NEURIPS2019_R2D2}               &\textbf{0.69}&0.91&0.95 &0.74&0.91&0.96 &\textbf{0.64}&\textbf{0.92}&0.94    &2.0K  \\
 &VGG19  \cite{simonyan2014very} &0.56&0.86&0.92 &0.60&0.89&0.95 & 0.51&0.84&0.89    &13.3K  \\
 &ResNet50 \cite{he2016deep}                    &0.52  &0.85  &0.92                 &0.57  &0.87  &0.94                 
 &0.47  &0.83  &0.90                 &13.7K                    \\
&ResNeXt50 \cite{xie2017aggregated}                    &0.50  &0.85  &0.92                 &0.55  &0.88  &0.94                 
&0.46  &0.83  &0.90                 &12.4K                    \\
&MobileV3 \cite{howard2019searching}                    &0.50  &0.82  &0.89                 &0.55  &0.84  &0.90                 
&0.45  &0.81  &0.88                 &12.8K                    \\
&EfficientNetV2 \cite{tan2021efficientnetv2}                    &0.50  &0.84  &0.92                 &0.56  &0.87  &0.93                 
&0.44  &0.82  &0.90                 &13.9K                    \\

\hline
 \multirow{3}{2.4cm}{\centering Semantic \\ 
 Segmentation}& FCN\cite{long2015fully}      &0.46& 0.78 & 0.85                &0.49  &0.78  &0.85                 &0.44  &0.77  &0.85                 &10.9K                    \\
 
&DeepLabV3 \cite{chen2017rethinking}                    &0.37  &0.62  &0.70                 &0.47  &0.76  &0.84                 
&0.26  &0.50  &0.57                 &4.4K                    \\

&Swin Tranformer\cite{liu2021swin}  &0.53  &0.80  &0.88                 &0.60  &0.81  &0.89                 &0.46  &0.79  &0.87                 &4.7K                    \\
\hline
\multirow{3}{2.4cm}{\centering Stereo\\Matching}&CREStereo\cite{li2022practical}     &0.51& 0.73 & 0.75                &0.57  &0.78  &0.81                 & 0.46 &0.68  & 0.70                &\textbf{23.3K}                    \\
&PSMNet\cite{chang2018pyramid}     &0.31& 0.51 & 0.56  &0.38  &0.58  &0.63            &0.24  &0.44  &0.49                                  &4.8K                    \\
&DeepPruner\cite{duggal2019deeppruner}     &0.31& 0.50 & 0.55  &0.37  &0.57  &0.62            &0.24  &0.44  &0.48                                  &5.2K                    \\

\hline
\multirow{1}{2.4cm}{\centering-}&Average Pooling     &0.60& 0.85 & 0.88                &0.66  &0.88  &0.92                 & 0.53 &0.81  & 0.84                &17.4K                    \\

\bottomrule
\end{tabular}

    \label{tab:exp_compare}
    \vspace{-0.6em}
\end{table*}

\begin{table*}[t!]
\vspace{+0.5em}
    \fontsize{7.5}{9.0}\selectfont
    \centering
    \caption{\textbf{Homography estimation results on the HPatches dataset.}
    }
    \vspace{-1em}
    \begin{tabular}{c c c c | c c c| c c c }
    \toprule
	\multirow{2}{*}{Method} & \multicolumn{3}{c|}{Overall} & \multicolumn{3}{c|}{Illumination} & \multicolumn{3}{c}{Viewpoint} \\

	 \cline{2-10}& @1px&@3px&@5px& @1px&@3px&@5px&@1px&@3px&@5px \\
  
	\midrule

    SIFT\cite{SIFT}+NNS  & 0.42&0.74&0.85 & 0.54&0.86&0.93 & 0.30&0.54&0.71   \\
    
    SuperPoint~\cite{detone2018superpoint}+NNS & 0.46&0.78&0.85 & 0.57&0.92&0.97 & 0.35&0.65&0.74   \\
    
    D2Net~\cite{dusmanu2019d2}+NNS & 0.38&0.71&0.82 & 0.66&0.95&0.98 & 0.12&0.49&0.67  \\
 
    R2D2~\cite{NEURIPS2019_R2D2}+NNS & 0.47&0.77&0.82 & 0.63&0.93&0.98 & 0.32&0.64&0.70   \\
 
 
 
    SuperPoint~\cite{detone2018superpoint}+SuperGlue~\cite{sarlin2020superglue} &  0.51&0.82&0.89 & 0.60&0.92&0.98 &0.42&0.71&0.81   \\
    
    SuperPoint~\cite{detone2018superpoint}+CAPS~\cite{wang2020learning}  & 0.49&0.79&0.86 & 0.62&0.93&0.98  & 0.36&0.65&0.75   \\

    SuperPoint~\cite{detone2018superpoint}+ClusterGNN~\cite{shi2022clustergnn} & 0.52&0.84&0.90 & 0.61&0.93&0.98  &  0.44&0.74&0.81   \\
    
    SIFT+CAPS~\cite{wang2020learning} & 0.36&0.77&0.85 & 0.48&0.89&0.95 & 0.26&0.65&0.76  \\
    
    Patch2Pix~\cite{zhou2021patch2pix}  & 0.50&0.79&0.87  & 0.71&0.95&0.98 & 0.30&0.64&0.76  \\

    MatchFormer &  0.55&0.81&0.87  &  \textbf{0.75}&0.95&0.98 & 0.37&0.68&0.78   \\

    LoFTR & \textbf{0.63}&\textbf{0.91}&\textbf{0.93} & 0.68&0.95&0.96  &  \textbf{0.59}&\textbf{0.86}&\textbf{0.91} \\
    
    \midrule
    
    DFM~\cite{dfm} &  0.41&0.74&0.85 & 0.63&0.91&0.97 & 0.21&0.59&0.74 \\
    
    Ours & 0.55&0.84&0.90 & 0.73&\textbf{0.95}&\textbf{0.98} & 0.38&0.73&0.81 \\

    \bottomrule	
\end{tabular}

    \label{tab:exp_homography}
    \vspace{-1.8em}
\end{table*}

\subsection{Implementation Details}

In our experiments, we employ ResNet18, pre-trained on the ImageNet dataset, as the default deep feature extractor. For the feature description, we use the patch descriptor from the distilled R2D2. We only incorporate a ratio test at the descriptor layer to correct errors from previous stages. Similar to the setup used in \cite{dfm}, two ratio tests (thresholds: 0.95 and 0.60) are applied. Our model is trained on the same datasets used to train R2D2. During the training process, we utilize the Adam optimizer with an initial learning rate of $1 \times 10^{-4}$ and a batch size of 2. The training process converges after 10 epochs on an NVIDIA GTX1650 GPU. The student model contains 292K parameters, compared to the teacher model, which has 486K parameters. This reduction in parameters demonstrates the efficiency gains achieved through our patch descriptor distillation strategy.

We incorporate a patch descriptor that allows the proposed method to be compatible with the most popular backbone networks. It is worth noting that our method is not directly compatible with the Swin Transformer architecture \cite{liu2021swin}, which initially applies a $4\times$ downsampling, resulting in a feature map with a maximum resolution equivalent to only a quarter of the input image. To ensure compatibility with Swin Transformer, we up-sample the points directly to match the required resolution. 

\subsection{Image Matching Performance Evaluation}
\label{sec:experiments_Image_Matching_Performance}
Table \ref{tab:exp_MMA} demonstrates that our proposed GCM outperforms all other existing algorithms in terms of overall and viewpoint MMA on the HPatches dataset. It achieves results comparable to LoFTR and better results than R2D2. 

Furthermore, the results presented in Table \ref{tab:exp_compare} confirm the compatibility of our algorithm with various backbone networks that are pre-trained for a wide range of computer vision tasks, including image classification, semantic segmentation, and stereo matching. Additionally, we propose an alternative approach that does not rely on a visual backbone network, by substituting the feature maps in the hierarchical refinement strategy with descriptors average pooling to different scales. This method is represented as Average Pooling in Table. \ref{tab:exp_compare}. The unsatisfactory results achieved with stereo-matching backbone networks are somewhat unexpected, and it is possible that these networks, primarily designed for 1D dense search, may not be well-suited for solving 2D search problems that are more relevant to correspondence matching tasks. This analysis provides valuable insights into the performance variation observed with different types of backbone networks and highlights the need for specialized feature extraction and matching strategies in correspondence matching applications. 
Considering the speed advantage of the ResNet18 model, we choose ResNet18 as the backbone for experiments.
It is also worth noting that the difference in performance between using the original R2D2 backbone and a lightweight distilled R2D2 backbone is minimal. This demonstrates the effectiveness of the distillation strategy in reducing computational complexity without sacrificing performance.
\begin{table}[!t]
    \fontsize{7.5}{9.0}\selectfont
    \centering
    \caption{\textbf{Pose estimation performance evaluation on the MegaDepth dataset.} These results denote the percentages of correctly estimated poses with pose errors below 5/10/20$^\circ$, respectively.
    }
    \vspace{-1em}
    \begin{tabular}{l c c c c}
    \toprule
	\multirow{2}{*}{Method} & \multicolumn{3}{c}{Pose Estimation AUC} & \multirow{2}{*}{Precision}\\
	  
   \cline{2-4}& @5$^\circ$ & @10$^\circ$ & @20$^\circ$ & \\
    \midrule
    SIFT+NNS & 16.70 & 28.15 & 41.84 & 35.53 \\
    D2-Net & 4.44 & 8.27 & 14.17 & 48.58 \\
    
    SuperPoint+NNS & 29.01 & 44.74 & 58.92 & 57.43 \\

    R2D2+NNS  & 41.15 & 58.88 & 72.84 & 81.51 \\
    
    SuperPoint+SuperGlue & 46.10 & 63.82 & 77.68 & \textbf{99.66} \\

    Patch2Pix & 39.70 & 55.06 & 67.77 & 80.17 \\
    
    LoFTR & \textbf{52.42} & \textbf{69.26} & \textbf{81.41} & 96.78 \\

    \midrule

    DFM & 25.91 & 41.70 & 56.19 & 91.92 \\
    
    Ours & 34.61 & 52.24 & 67.10 & 87.41\\

    \bottomrule	
\end{tabular}
    \label{tab:exp_MegaDepth}
    \vspace{-1.2em}
\end{table}
\subsection{Homography Estimation Performance Evaluation}
In Table \ref{tab:exp_homography}, the accuracy metrics reported for overall, illumination, and viewpoint matching at various thresholds provide a comprehensive performance evaluation of homography estimation for both our method and the compared algorithms. Notably, while GCM achieves the second-best performance across all these categories, it outperforms the majority of data-driven approaches that are trained using fully supervised correspondence ground truth. 

\begin{table}[t!]
    \fontsize{7.5}{9.0}\selectfont
    \centering
    \caption{\textbf{Pose estimation performance evaluation on the ScanNet dataset.} ``$\dagger$" denotes the models trained on the outdoor datasets and evaluated on the ScanNet dataset (indoor).
    }
    \vspace{-1em}
    \begin{tabular}{l c c c c}
    \toprule
	\multirow{2}{*}{Method} & \multicolumn{3}{c}{Pose Estimation AUC} & \multirow{2}{*}{Precision}\\
	 \cline{2-4}& @5$^\circ$ & @10$^\circ$ & @20$^\circ$ &\\
	\midrule

    SIFT+NNS & 5.83 & 13.06 & 22.47 & 40.30 \\
    SuperPoint+NNS& 9.43 & 21.53 & 36.40 & 50.40 \\
    Patch2Pix$^{\dagger}$ & 9.57 & 21.22 & 34.56 & 50.59 \\
    R2D2$^{\dagger}$& 6.82 & 16.37 & 28.02 & 46.51 \\
    SuperPoint+SuperGlue& 16.24 & 34.13 & 52.88 & \textbf{84.33} \\
    SuperPoint+SuperGlue$^{\dagger}$& 15.68 & 32.66 & 49.87 & 80.45 \\
    LoFTR& \textbf{20.27} & \textbf{39.63} & \textbf{57.47} & 83.92 \\
    LoFTR$^{\dagger}$& 17.57  & 34.46 & 51.88 & 70.16 \\   
    \midrule
    DFM& 3.73 & 9.76 & 18.94 & 74.77 \\
    Ours$^{\dagger}$& 11.03 & 23.73 & 38.93 & 67.63 \\

    \bottomrule	
\end{tabular}
    \label{tab:exp_ScanNet}
    \vspace{-1.2em}
\end{table}

\subsection{Pose Estimation Performance Evaluation} 

Tables \ref{tab:exp_MegaDepth} and \ref{tab:exp_ScanNet} provide the performance evaluations for pose estimation in outdoor and indoor environments, respectively. The MegaDepth dataset presents a significant challenge due to the necessity of matching under extreme viewpoint changes and addressing issues related to high texture repetition. Despite utilizing a non-specialized network for deep feature extraction, our method consistently demonstrates impressive performance on demanding test data. While it may not reach the state-of-the-art performance level achieved by fully supervised methods in pose estimation, our approach represents a notable improvement over the baseline plug-and-play algorithm DFM.
Additionally, the experimental results on the ScanNet dataset demonstrate that our proposed GCM achieves performance levels comparable to fully supervised methods and outperforms the baseline algorithm DFM across most threshold values.
\subsection{Abaltion Studies}
\begin{table}[!t]
    \fontsize{7.5}{9.0}\selectfont
    \centering
    \caption{ Ablation Study of each component on Hpatches dataset.
    }
    \vspace{-1em}
    
\begin{tabular}{
    c
    c 
    c  
    >{\centering\arraybackslash}p{0.9cm} 
    >{\centering\arraybackslash}p{0.9cm} 
    >{\centering\arraybackslash}p{0.9cm}}
    \toprule
    \multirow{2}{*}{FHR} & \multirow{2}{*}{Descriptor} & \multirow{2}{*}{Distillation} & \multicolumn{3}{c}{Homography Estimation Accuracy} \\
    \cline{4-6}
    & & & @1px & @3px & @5px \\
    \midrule
      &  &  & 0.37 & 0.68 & 0.80 \\
      & \ding{51} &  & 0.41& 0.72 & 0.82\\
      & \ding{51} &\ding{51}  & 0.42 & 0.72 & 0.83 \\
     \ding{51}&  &  & 0.38 & 0.71 & 0.83 \\
     \ding{51}&\ding{51}&& \textbf{0.45} & 0.75 & 0.84 \\
     \ding{51}&\ding{51}&\ding{51}& 0.44 & \textbf{0.76} & \textbf{0.85} \\

    \bottomrule
\end{tabular}
    \label{tab:hpatches_ablation_homo}
    \vspace{-1.2em}
\end{table}
We conduct an ablation study to determine the individual and combined effects of FHR, patch descriptor, and descriptor distillation on the homography estimation accuracy across the overall Hpatches dataset, with the setup as referenced in Sec. \ref{sec:experiments_Image_Matching_Performance}. The results are presented in Table \ref{tab:hpatches_ablation_homo}. VGG19 is utilized as the backbone network and the baseline model (the first row) is designated as DFM, with random sample consensus (RANSAC) for homography matrix estimation to ensure experimental consistency. Our key insights: 
\begin{itemize}
    \item Patch descriptors play a crucial role in improving accuracy, in the configurations where it is employed (the second and fifth rows). These results underscore its essential contribution to baseline performance enhancement and integration with FHR for further improvements.
    
    \item Descriptor distillation significantly refines patch descriptor quality, leading to more accurate homography estimation. This is evident in the configurations incorporating descriptor distillation  (the third and sixth rows).
    
    \item FHR contributes to a notable increase in accuracy (from the fourth to the sixth rows). When applied in combination with patch descriptor and distillation, FHR significantly elevates the model's overall performance.
\end{itemize}
\section{Conclusion}
In summary, this paper introduced three significant technical contributions to address the limitations present in DFM: (1) a flexible hierarchical refinement strategy that eliminates the need for a pre-defined threshold, initially utilized in the hierarchical refinement process of DFM; (2) the incorporation of a patch descriptor that extends the applicability of DFM to accommodate a wide range of backbone networks, pre-trained across diverse robot perception tasks, such as semantic segmentation and stereo matching; (3) a novel patch descriptor distillation strategy, which further reduces the computational complexity of correspondence matching and enhances the practical applicability of our method in real-world robotics applications. Through performance evaluations including image matching, homography estimation, and pose estimation, the effectiveness of our proposed algorithm is validated. Particularly noteworthy is that our method achieves state-of-the-art overall mean matching accuracy, outperforming both conventional hand-crafted approaches and data-driven methods trained via fully supervised learning. We are confident that our method can be readily integrated into a variety of downstream real-world robotics applications.

\clearpage
\bibliographystyle{ieeetr}
\bibliography{references}

\end{document}